\documentclass[letterpaper]{article} 
\usepackage{aaai25}  
\usepackage{times}  
\usepackage{helvet}  
\usepackage{courier}  
\usepackage[hyphens]{url}  
\usepackage{graphicx} 
\usepackage{natbib}  
\usepackage{caption} 
\usepackage{algorithm}
\usepackage{algorithmic}

\usepackage{makecell}
\usepackage{anyfontsize}
\usepackage{amsmath}
\usepackage{booktabs}
\usepackage{amssymb}
\usepackage{subcaption}

\usepackage{newfloat}
\usepackage{listings}
\DeclareCaptionStyle{ruled}{labelfont=normalfont,labelsep=colon,strut=off} 
\floatstyle{ruled}
\newfloat{listing}{tb}{lst}{}
\floatname{listing}{Listing}
\title{Bridge Diffusion Model: Bridge Chinese Text-to-Image Diffusion Model with English Communities}
\author{
    Shanyuan Liu, Bo Cheng, Yuhang Ma, Liebucha Wu \\Ao Ma, Xiaoyu Wu, Dawei Leng\thanks{Corresponding authors.}, Yuhui Yin
}
\affiliations{
    360 AI Research \\
    \{liushanyuan, chengbo1, mayuhang, wuliebucha\}@360.cn \\
    \{maao, wuxiaoyu1, lengdawei, yinyuhui\}@360.cn
}

\usepackage{bibentry}

\begin{document}

\maketitle

\begin{figure*}[t]
\centering
\includegraphics[width=1\linewidth]{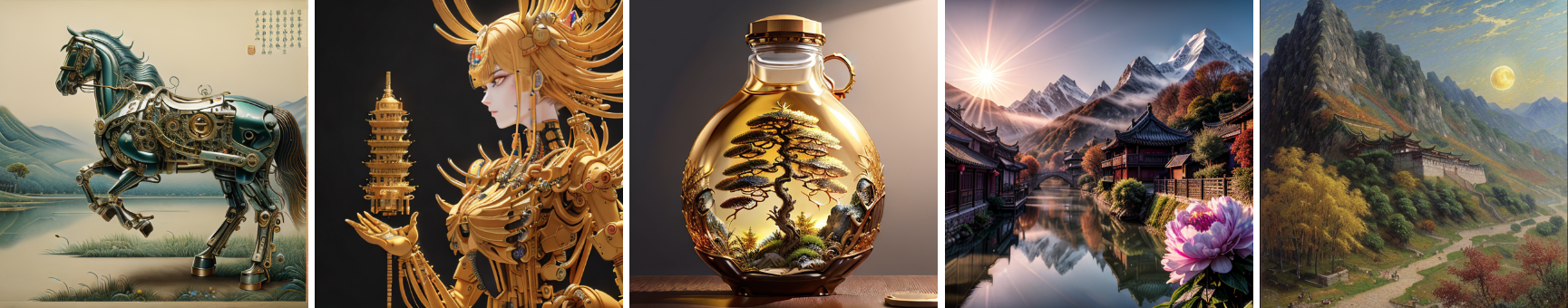} 
\caption{Selected samples generated by our BDM model, with Chinese prompts and different English plugins from the Stable Diffusion community.}
\label{fig:showcase}
\end{figure*}

\begin{abstract}
Text-to-Image generation (TTI) technologies are advancing rapidly, especially in the English language communities. However, apart from the user input language barrier problem, English-native TTI models inherently carry biases from their English world centric training data, which creates a dilemma for development of other language-native TTI models. One common choice is to fine-tune the English-native TTI model with translated samples. It falls short of fully addressing the model bias problem. Alternatively, training non-English language native models from scratch can effectively resolve the English world bias, but model trained this way would diverge from the English TTI communities, thus not able to utilize the strides continuously gaining in the English TTI communities any more. To build Chinese TTI model meanwhile keep compatibility with the English TTI communities, we propose a novel model structure referred as "Bridge Diffusion Model" (BDM). The proposed BDM employs a backbone-branch network structure to learn the Chinese semantics while keep the latent space compatible with the English-native TTI backbone, in an end-to-end manner. The unique advantages of the proposed BDM are that it's not only adept at generating images that precisely depict Chinese semantics, but also compatible with various English-native TTI plugins, such as different checkpoints, LoRA, ControlNet, Dreambooth, and Textual Inversion, \textit{etc}. Moreover, BDM can concurrently generate content seamlessly combining both Chinese-native and English-native semantics within a single image, fostering cultural interaction.
\end{abstract}

\begin{links}
\link{Code}{https://github.com/360CVGroup/Bridge_Diffusion_Model}
\link{Model}{https://huggingface.co/qihoo360/BDM1.0}
\end{links}

\section{Introduction}
The latest advancements in diffusion models have significantly transformed the process of text-to-image generation. With extensive training data and model parameters, diffusion models\cite{Ho_Jain_Abbeel_2020} can now vividly depict visual scenes based on written prompts, allowing users to effortlessly generate beautiful images using natural language. Among these models, Stable Diffusion\cite{Rombach_Blattmann_Lorenz_Esser_Ommer_2022} has emerged as a widely adopted and community-driven technology. It maps images from pixel space to latent space, resulting in impressive image quality while significantly reducing memory usage. The SD community has made important progress, introducing various LoRA\cite{Hu_Shen_Wallis_Allen-Zhu_Li_Wang_Chen_2021} models and checkpoints. These additions enhance the base model's capabilities, enabling it to generate more refined or personalized content within specific domains. It's worth noting that the rapid progress of these technologies is mainly centered around English-language communities.

However, current models exhibit language-related biases. As pointed out by \cite{DBLP:journals/corr/abs-2303-11408,miller2023ai}, Text-to-Image (TTI) models designed for English speakers, such as DALL-E 2\cite{DBLP:journals/corr/abs-2204-06125}, Stable Diffusion\cite{Rombach_Blattmann_Lorenz_Esser_Ommer_2022} versions 1.4 and 2, tend to disproportionately emphasize characteristics associated with white individuals and males. These language-related biases are inherent and widespread in current TTI models, primarily because they are predominantly trained on data from the English-speaking world, as exemplified by the commonly used LAION dataset\cite{DBLP:conf/nips/SchuhmannBVGWCC22}. Consequently, there is an over-representation of English-speaking figures and an inadequate representation of non-English-speaking counterparts. 

The primary focus of this work is to design a TTI model that is compatible with multiple languages. By "compatible", we mean not only supporting non-English prompt input but also capable of generating images that align with common sense in non-English languages. While translation-based methods are straightforward and cost-effective, they can only address the non-English input capability, leaving the inherent model bias untouched. Another approach involves alignment-based strategies, which align the embedding space of different language text encoders with parallel translation text corpus. However, this method is essentially another form of "translation". Taiyi-Stable-Diffusion-1B-Chinese-EN-v0.1\cite{fengshenbang} took this route, fine-tuning the TTI model with Chinese-native data based on an aligned text encoder. This allows the English-native model to incorporate Chinese-native language semantics at a low cost while maintaining some level of compatibility between the English and Chinese TTI communities, although achieving the right balance is challenging. However, when tasked with capturing intricate nuances of native language semantics or language-specific concepts, the effectiveness of this approach may be notably limited. To address the inherent bias in English-native models, the most radical approach is to train a TTI model from scratch with non-English native data. For example, ERNIE-ViLG 2.0\cite{feng2023ernie} and Wukong-Huahua\cite{gu2022wukong} are trained with Chinese native data and can generate high-quality images consistent with Chinese language semantics. However, a fundamental challenge with this approach is that it loses compatibility with its ancestral English-native models, preventing the direct utilization of progress from the English-native TTI communities. This could lead to community isolation and development stagnation for the Chinese-native TTI community in the long run.

We propose a new diffusion network structure referred to as the "Bridge Diffusion Model" (BDM) to address the previously mentioned challenge. The unique advantages of the proposed BDM lie in its ability to precisely generate images following non-English language native semantics while also being compatible with English-native TTI communities. Existing techniques from English-native TTI communities, such as different checkpoints, LoRA, ControlNet, Dreambooth, and Textual Inversion, can all be directly applied in BDM. BDM effectively addresses the language-related bias in TTI models while maintaining interoperability between non-English language native TTI communities and English-native communities. This is where the name "Bridge" originates. In this context, we specifically focus on implementing Chinese language native TTI realization, and the method could be applicable for any other non-English language native TTI model.

Our approach involves employing a backbone-branch network architecture, similar to ControlNet\cite{zhang2023adding}, as illustrated in Fig.\ref{fig:BDM structure}. The backbone remains frozen during training and can be sourced from any pre-trained diffusion TTI model. In our current implementation, we utilize Stable Diffusion 1.5\cite{Rombach_Blattmann_Lorenz_Esser_Ommer_2022}. The branch functions as a module for injecting language-native semantics, and its parameters are trained using language-native text-image pairs. In contrast to ControlNet, BDM's branch incorporates a Chinese-native CLIP\cite{chinese-clip} as the text encoder, responsible for processing the non-English language-native text prompts. The English-native text encoder is fed with an empty constant string ("") in our implementation.

For model inference, language-native prompts will be processed through the Chinese text encoder from BDM's branch part. Simultaneously, we can still input an empty constant string ("") into the English text encoder. Since BDM incorporates an entire English-native TTI model as its backbone, existing techniques like LoRA\cite{Hu_Shen_Wallis_Allen-Zhu_Li_Wang_Chen_2021}, ControlNet\cite{zhang2023adding}, Dreambooth\cite{ruiz2023dreambooth}, Textual Inversion\cite{DBLP:conf/iclr/GalAAPBCC23}, and even various style fine-tuned checkpoints from English TTI communities (such as Civitai\cite{civitai}, Stable Diffusion Online\cite{stablediffusionweb}, to name a few) can be seamlessly applied to BDM with minimal cost.

In summary, the primary contributions of this work are as follows: (1) We propose a backbone-branch network architecture referred as BDM. By integrating an English TTI backbone with a Chinese-native semantics injection branch, BDM is able to solve TTI model's English-native bias meanwhile keep the compatibility with its ancestral English-native communities. (2) We proposed a tailored training strategy for BDM. By utilizing the latent space corresponding to the empty constant string ("") of the English backbone, we aligned the Chinese-native latent space with the English latent space, enabling BDM to incorporate English communities plugins. (3) With thorough experiments, we verify that the BDM trained with Chinese-native data can utilize techniques developed for English-native TTI models, such as LoRA, Dreambooth, Textual Inversion, ControlNet, \textit{etc}. Thus bridge the interoperation between non-English and English-native TTI communities.

\section{Related Work}
\label{sec:Related Work}

In recent years, the field of TTI generation has experienced remarkable growth. Early endeavors employed Generative Adversarial Networks \cite{Goodfellow_Pouget-Abadie_Mirza_Xu_Warde-Farley_Ozair_Courville_Bengio_2017} to produce images from textual descriptions. This was achieved by harnessing the adversarial interplay between a generator and a discriminator to yield lifelike images. With the resounding success of the transformer architecture, a subset of research shifted the generation task towards a sequence-to-sequence paradigm, training generators in an autoregressive manner. Noteworthy instances encompass ERNIE-ViLG \cite{DBLP:journals/corr/abs-2112-15283}, DALL-E \cite{Ramesh_Pavlov_Goh_Gray_Voss_Radford_Chen_Sutskever_2021}, and Parti \cite{DBLP:journals/tmlr/YuXKLBWVKYAHHPLZBW22}. More recently, diffusion models have garnered acclaim for achieving cutting-edge outcomes in this domain \cite{DBLP:journals/corr/abs-2204-06125,Rombach_Blattmann_Lorenz_Esser_Ommer_2022,saharia2022photorealistic}. These methods have continually refreshed the metrics in this field by iteratively injecting text conditions during the denoising process. Evident progress can be seen in works like LDM \cite{Rombach_Blattmann_Lorenz_Esser_Ommer_2022}, DALL-E 2 \cite{DBLP:journals/corr/abs-2204-06125}, and Imagen \cite{saharia2022photorealistic}. Recently, diffusion models based on the DIT have been emerging in abundance, with some even achieving state-of-the-art performance, such as SD3.5\cite{SD3.5} and Flux\cite{FLUX}. The primary aim of our research is to leverage the backbone-branch architecture to empower the network in generating images that possess a profound grasp of native language semantic understanding.

Moreover, distinct research endeavors have concentrated on directing the output of diffusion models to finely manage image content and style. Techniques like Dreambooth\cite{ruiz2023dreambooth} and Textual Inversion\cite{DBLP:conf/iclr/GalAAPBCC23} bestow precise control over the attributes of generated images, accomplishing objectives analogous to reference images. LoRA\cite{Hu_Shen_Wallis_Allen-Zhu_Li_Wang_Chen_2021} facilitates lightweight fine-tuning of the base model to achieve specified objectives or styles. ControlNet\cite{zhang2023adding}, T2I-Adapter\cite{mou2023t2i} and HiCo\cite{DBLP:journals/corr/abs-2410-14324} facilitate meticulous control over diverse conditions by incorporating branches into pretrained base models. Within our study, our focus is centered around seamlessly integrating BDM with the control plugins from the aforementioned English communities.

In addition, research has also been devoted to enhancing the multilingual capability of TTI models to support non-English input captions. For example, Altclip\cite{DBLP:conf/acl/ChenLZYW23} extended the text encoder of diffusion models using the pretrained multilingual text encoder XLM-R. ERNIE-ViLG 2.0\cite{feng2023ernie} embarked on training a Chinese diffusion model from scratch using Chinese image-text pairs. \cite{DBLP:conf/acl/LiCRVK23} alleviated the language barrier by translating English captions into other languages through a neural machine translation system. Although they improved the support for multilingual text inputs, they did not achieve the integration of native language and English communities, while our focus is on incorporating native language models into the English communities.

\section{Preliminary}
\label{sec:Preliminary}
\subsection{Diffusion model}
Diffusion models \cite{Ho_Jain_Abbeel_2020,DBLP:conf/icml/Sohl-DicksteinW15,Song_Ermon_2019,Vincent_2011} are a class of generative models that have gained significant attention recently due to their advancements in image generation \cite{Dhariwal_Nichol_2021,kawar2022enhancing,Song_Sohl-Dickstein_Kingma_Kumar_Ermon_Poole_2020,Vahdat_Kreis_Kautz_2021}, leading to the latest technological developments in several downstream applications. These applications include image restoration \cite{DBLP:conf/nips/KawarEES22,Saharia_Chan_Chang_Lee_Ho_Salimans_Fleet_Norouzi_2022}, adversarial purification \cite{DBLP:journals/corr/abs-2207-08089,DBLP:conf/icml/NieGHXVA22}, image compression \cite{DBLP:journals/corr/abs-2206-08889}, image classification \cite{DBLP:journals/corr/abs-2110-00473}, among others in various fields \cite{DBLP:journals/corr/abs-2207-03442,DBLP:journals/corr/abs-2209-11888,Popov_Vovk_Gogoryan_Sadekova_Kudinov_2021,Sasaki_Willcocks_Breckon_2021}. 

The fundamental concept of DDPM\cite{Ho_Jain_Abbeel_2020} involves iteratively applying diagonal Gaussian noise to the initial data sample (denoted as $x$) and, after $T$ steps, transforming it into an isotropic Gaussian distribution
\begin{equation}
x_t=\sqrt{\alpha_t} x_{t-1}+\sqrt{1-\alpha_t} \epsilon_t, \quad t \in\{1, \ldots, T\}
\end{equation}
Here, the sequence $\left\{x_t\right\}$ starts from $x_0=x$ and ends with $x_T \sim \mathcal{N}(0, I)$, where each step adds noise $\epsilon_t \sim \mathcal{N}(0, I)$, and $\left\{\alpha_t\right\}_{1 \ldots T}$  represents the predefined schedule . The training objective of this network is straightforward denoising, aiming to make $\epsilon_\theta\left(x_t, t\right) \approx \epsilon_t$. This leads to a highly consistent learned image distribution with the target distribution, resulting in excellent generative performance.

\section{Method}
\label{sec:Method}
\begin{figure}[t]
\centering
\includegraphics[width=1\columnwidth]{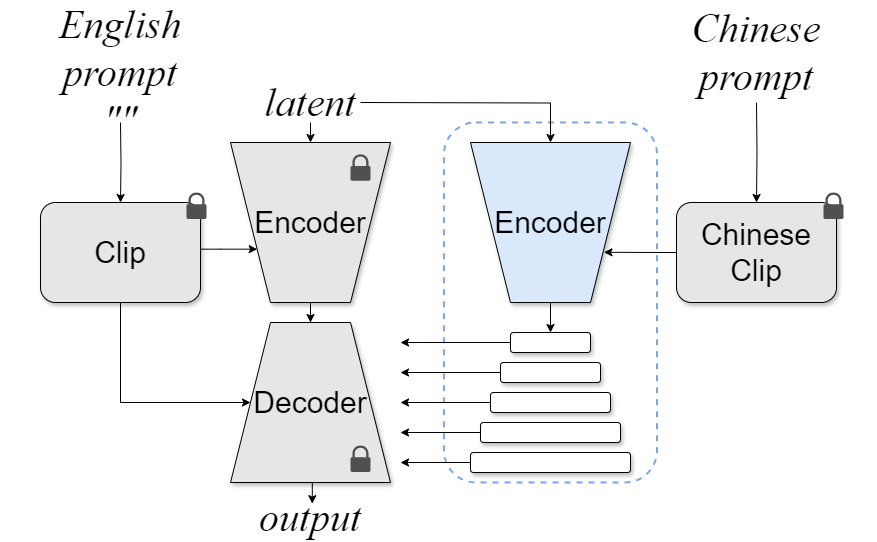} 
\caption{BDM structure.}
\label{fig:BDM structure}
\end{figure}

The overall framework of BDM is shown in the Fig.\ref{fig:BDM structure}. It adopts a backbone-branch architecture similar to ControlNet\cite{zhang2023adding} and the backbone is responsible for providing the same latent space as the English communities, while the branch is responsible for finding appropriate offsets in the latent space, allowing native language semantic features to be injected into the English latent space.

\subsection{backbone-branch architecture}

For BDM, we embrace a backbone-branch architecture akin to ControlNet\cite{zhang2023adding}. The backbone employs Stable Diffusion 1.5\cite{Rombach_Blattmann_Lorenz_Esser_Ommer_2022}, and the branch consists of a learnable parameter replica derived from the backbone, with convolutional layers responsible for processing conditional image inputs omitted. In addition, the backbone and branch process the same latent space features. For the backbone, text encoding is accomplished via OpenAI CLIP\cite{DBLP:conf/icml/RadfordKHRGASAM21}, while the branch employs Chinese CLIP\cite{chinese-clip} for text encoding. This framework harmoniously marries the latent space and generation capabilities of BDM with the English model. Simultaneously, the branch accommodates native language semantics, thereby facilitating the creation of a native language model that seamlessly aligns with the English communities' dynamics.
\subsection{training strategy}
Image diffusion models learn the process of progressively denoising images to generate new samples. The denoising can take place either in the pixel space or a latent space that is encoded from the training data. Stable Diffusion, for instance, employs latent images as the domain for training. In this context, the terms "image", "pixel", and "denoising" all refer to their corresponding concepts in the "perceptual latent space" \cite{Rombach_Blattmann_Lorenz_Esser_Ommer_2022}.

The diffusion algorithms start with an initial image $z_0$, and then iteratively add noise to produce a series of noisy images $z_t$, where $t$ indicates the number of times noise is added. As $t$ increases, the image approximates pure noise. In this context, image diffusion algorithms learn a network $\epsilon_\theta$ to predict the noise to be added to the noisy image $z_t$, given a set of conditions, including the time step $t$ ,text prompts for the English backbone $c_{ent}$ ,text prompts for the native language branch $c_{nlt}$

\begin{equation}
\begin{aligned}
    \mathcal{L} = &\mathbb{E}_{\boldsymbol{z}_0, t, \boldsymbol{c}_{ent},\boldsymbol{c}_{nlt},  \epsilon \sim \mathcal{N}(0,1)} \\ 
    &\left[ || \epsilon - \epsilon_\theta(z_t, t, \boldsymbol{c}_{ent},\boldsymbol{c}_{nlt}) ||_2^2 \right]
\end{aligned}
\end{equation}

The learning objective $\mathcal{L}$ represents the overall objective of the entire diffusion model and can be directly employed for training BDM.

Our first key finding here is that to successfully train the BDM model, it's important to only inject text prompt information through the non-English language branch, meanwhile leave the text input of the English backbone empty, otherwise the training would not converge. Reasons are two folds: on the one hand, the English backbone here is used only for latent space alignment, \textit{i.e.} for keeping BDM's compatibility with the English SD model, thus there's no need for text input of the English backbone; one the other hand, if the text input of the English backbone is non-empty, it would produce strong effect on the UNet denoising process, thus would significantly impede the effective injection of native language semantics through the branch network. Therefore, throughout the training process, we consistently set $c_{ent}$ as an empty string, ensuring a dependable and aligned latent space for the BDM. The BDM utilizes the latent space of English backbone to find the optimal shift in native language semantics within the English domain, akin to how English community plugins seek specific semantic shifts in the latent space of Stable Diffusion 1.5. It's worth noting that, with a shared latent space, English community plugins seamlessly integrate with each other, and the BDM, following the same principle and latent space, can also seamlessly integrate with these plugins, as verified in subsequent experiments.

\subsection{inference strategy}
\label{inference strategy}
Unlike the training strategy mentioned earlier, our second key finding is that, during inference, BDM exhibits additional versatility. By manipulating the text prompt input to different language branches, we can generate images that emphasize Chinese semantics, English semantics, or even a combination of both.

To start, we have the option to set both the prompt and negative prompt of the English backbone to empty strings. This practice aligns with the training phase, solely utilizing prompts from the native language branch to infuse semantics. We observed that using appropriate positive/negative prompts in the English backbone during the inference stage can enhance image quality or alter the visual style. Therefore, for better generation results or a more suitable style, both Chinese and English prompts, along with negative prompts, can be adjusted based on user preferences or requirements. Moreover, the weight of the native language branch can also be changed according to user needs.

BDM also has the ability to integrate various plugins from the English communities. Incorporating checkpoint or Dreambooth\cite{ruiz2023dreambooth} involves a direct switch from the backbone's Stable Diffusion 1.5 to the desired checkpoint/Dreambooth. When incorporating LoRA\cite{Hu_Shen_Wallis_Allen-Zhu_Li_Wang_Chen_2021}—be it style LoRA or object LoRA—embedding the LoRA model parameters into the backbone is feasible. If LoRA contains trigger words, it is crucial to input these trigger words into the English backbone. Similarly, when integrating with ControlNet\cite{zhang2023adding}, the ControlNet branch can seamlessly combine with the backbone, resulting in a dual-branch configuration comprising the BDM native language branch and the original ControlNet branch. Regarding Textual Inversion\cite{DBLP:conf/iclr/GalAAPBCC23} integration, Textual Inversion’s embedding can be directly loaded into the backbone’s prompt or negative prompt. The combination of these operations can be customized according to specific needs.

In contrast, when it comes to constructing native language models from scratch, it becomes imperative to retrain requisite plugins like LoRA due to the lack of compatibility with English-speaking communities. Nevertheless, BDM stands out by seamlessly incorporating plugins from English-speaking communities, providing an inherent advantage that facilitates effortless compatibility with these communities.

\section{Experiments}
\label{sec:Experiments}

\begin{figure}[t]
  \centering
  \begin{subfigure}{0.49\columnwidth}
    \includegraphics[width=\linewidth]{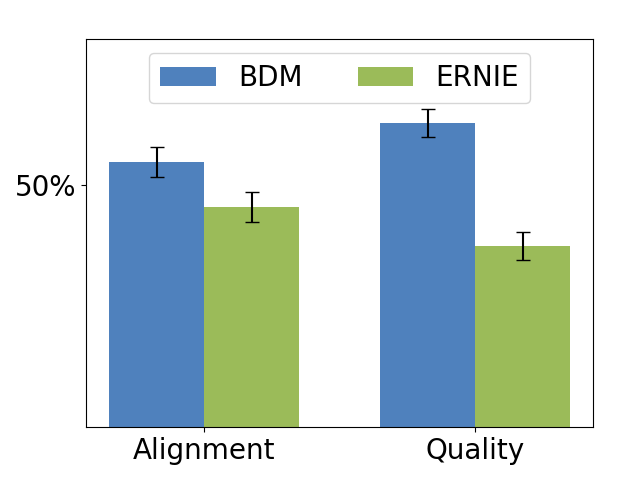}
    \caption{ERNIE-ViLG-2.0}
    \label{fig:ERNIE-ViLG-2.0}
  \end{subfigure}
  \begin{subfigure}{0.49\columnwidth}
    \includegraphics[width=\linewidth]{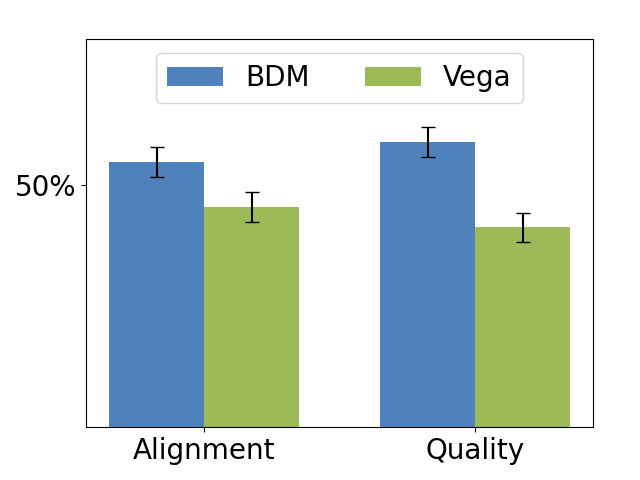}
    \caption{Vega}
    \label{fig:Vega}
  \end{subfigure}
    \caption{Comparison of BDM and ERNIE-ViLG-2.0 (referred to as ERNIE in the figure)/Vega on BDM-870 with human evaluation. In the human experiments, BDM utilized realisticVisionV51 as the English backbone to achieve enhanced generation results.}
  \label{fig:human}
  \vspace{-1em}
\end{figure}

\begin{table}[t]
  \centering
    \resizebox{.95\columnwidth}{!}{
    \begin{tabular}{@{}lcc@{}}
    \toprule
    \normalsize \rmfamily Model & \normalsize \rmfamily Data distribution & \normalsize \rmfamily FID $\downarrow$  \\
    \midrule
    \normalsize \rmfamily ERNIE-ViLG~\cite{DBLP:journals/corr/abs-2112-15283}                  & \normalsize \rmfamily English and Chinese      & \normalsize \rmfamily 14.70 \\
    \normalsize \rmfamily LDM~\cite{Rombach_Blattmann_Lorenz_Esser_Ommer_2022}                 & \normalsize \rmfamily English                  & \normalsize \rmfamily 12.61 \\
    \normalsize \rmfamily GLIDE~\cite{DBLP:conf/icml/NicholDRSMMSC22}                          & \normalsize \rmfamily English                  & \normalsize \rmfamily 12.24 \\
    \normalsize \rmfamily DALL-E~2~\cite{DBLP:journals/corr/abs-2204-06125}                    & \normalsize \rmfamily English                  & \normalsize \rmfamily 10.39 \\
    \normalsize \rmfamily Stable Diffusion~\cite{Rombach_Blattmann_Lorenz_Esser_Ommer_2022}    & \normalsize \rmfamily English                  & \normalsize \rmfamily \textbf{8.59} \\
    \normalsize \rmfamily eDiff-I~\cite{balaji2022ediffi}                                      & \normalsize \rmfamily English                  & \normalsize \rmfamily 6.95 \\
    \normalsize \rmfamily ERNIE-ViLG~2.0\cite{feng2023ernie}                                   & \normalsize \rmfamily English and Chinese      & \normalsize \rmfamily 6.75 \\
    \normalsize \rmfamily RAPHAEL~\cite{DBLP:journals/corr/abs-2305-18295}                     & \normalsize \rmfamily English                  & \normalsize \rmfamily 6.61 \\
    \midrule
    \normalsize \rmfamily BDM~ & \normalsize \rmfamily Chinese & \normalsize \rmfamily \textbf{9.93} \\
    \bottomrule
    \end{tabular}
    }
    \caption{Comparison of BDM and representative text-to-image generation models on MS-COCO $256 \times 256$ with zero-shot FID-30k. We use classifier-free guidance scale 3.7 and Chinese language branch weight 1.5 for our diffusion model. Here we choose SD1.5 as the English backbone for BDM}
  \label{tab:Comparison}
\end{table}

\begin{table}[t]
  \centering
    \resizebox{.95\columnwidth}{!}{
  \begin{tabular}{@{}cccccccccc@{}}
        \toprule
       \normalsize \rmfamily class & \multicolumn{2}{c}{\normalsize \rmfamily human}  & \multicolumn{2}{c}{\normalsize \rmfamily architecture}  \\
       \normalsize \rmfamily culture & \normalsize \rmfamily Chinese & \normalsize \rmfamily Caucasian & \normalsize \rmfamily Chinese & \normalsize \rmfamily Caucasian \\
        \midrule
        \normalsize \rmfamily BDM(RealV51) & \textbf{\normalsize \rmfamily 23.48} & \normalsize \rmfamily 22.05 & \textbf{\normalsize \rmfamily 28.59} & \normalsize \rmfamily 27.81 \\
        \normalsize \rmfamily RealV51 & \normalsize \rmfamily 22.86 & \textbf{\normalsize \rmfamily 23.24} & \normalsize \rmfamily 26.19 & \textbf{\normalsize \rmfamily 28.08} \\
        \bottomrule
    \end{tabular}
    }
    \resizebox{.95\columnwidth}{!}{
  \begin{tabular}{@{}cccccccccc@{}}
        \toprule
       \normalsize \rmfamily class & \multicolumn{2}{c}{\normalsize \rmfamily food}  &   \multicolumn{2}{c}{\normalsize \rmfamily festival} \\
       \normalsize \rmfamily culture & \normalsize \rmfamily Chinese & \normalsize \rmfamily Caucasian & \normalsize \rmfamily Chinese & \normalsize \rmfamily Caucasian \\
        \midrule
        \normalsize \rmfamily BDM(RealV51) & \textbf{\normalsize \rmfamily 24.77} & \normalsize \rmfamily 24.02 & \textbf{\normalsize \rmfamily 23.75} & \normalsize \rmfamily 20.77 \\
        \normalsize \rmfamily RealV51 & \normalsize \rmfamily 23.65 & \textbf{\normalsize \rmfamily 24.81} & \normalsize \rmfamily 19.20 & \textbf{\normalsize \rmfamily 24.03} \\
        \bottomrule
    \end{tabular}
    }
    \caption{CLIP score$\uparrow$ to measure Chinese cultural inclination}
    \label{tab:clip score}
    \vspace{-1em}
\end{table}


\begin{figure}[t]
\centering
\includegraphics[width=0.9\columnwidth]{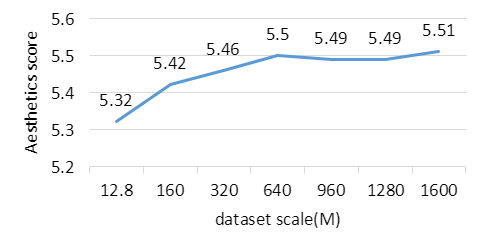}
\caption{Training data scale.}
\label{fig:dataset scale}
\end{figure}

\begin{figure}[t]
\centering
\includegraphics[width=0.9\columnwidth]{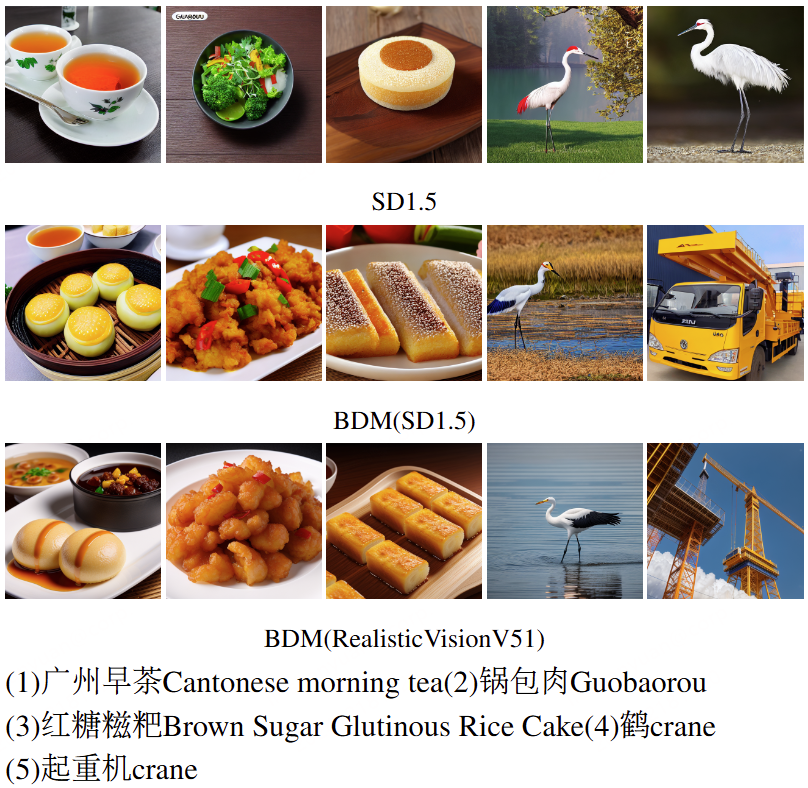} 
\caption{Native language semantics.}
\label{fig:native language semantic Checkpoint}
\vspace{-1em}
\end{figure}

\begin{figure}[t]
\centering
\includegraphics[width=0.9\columnwidth]{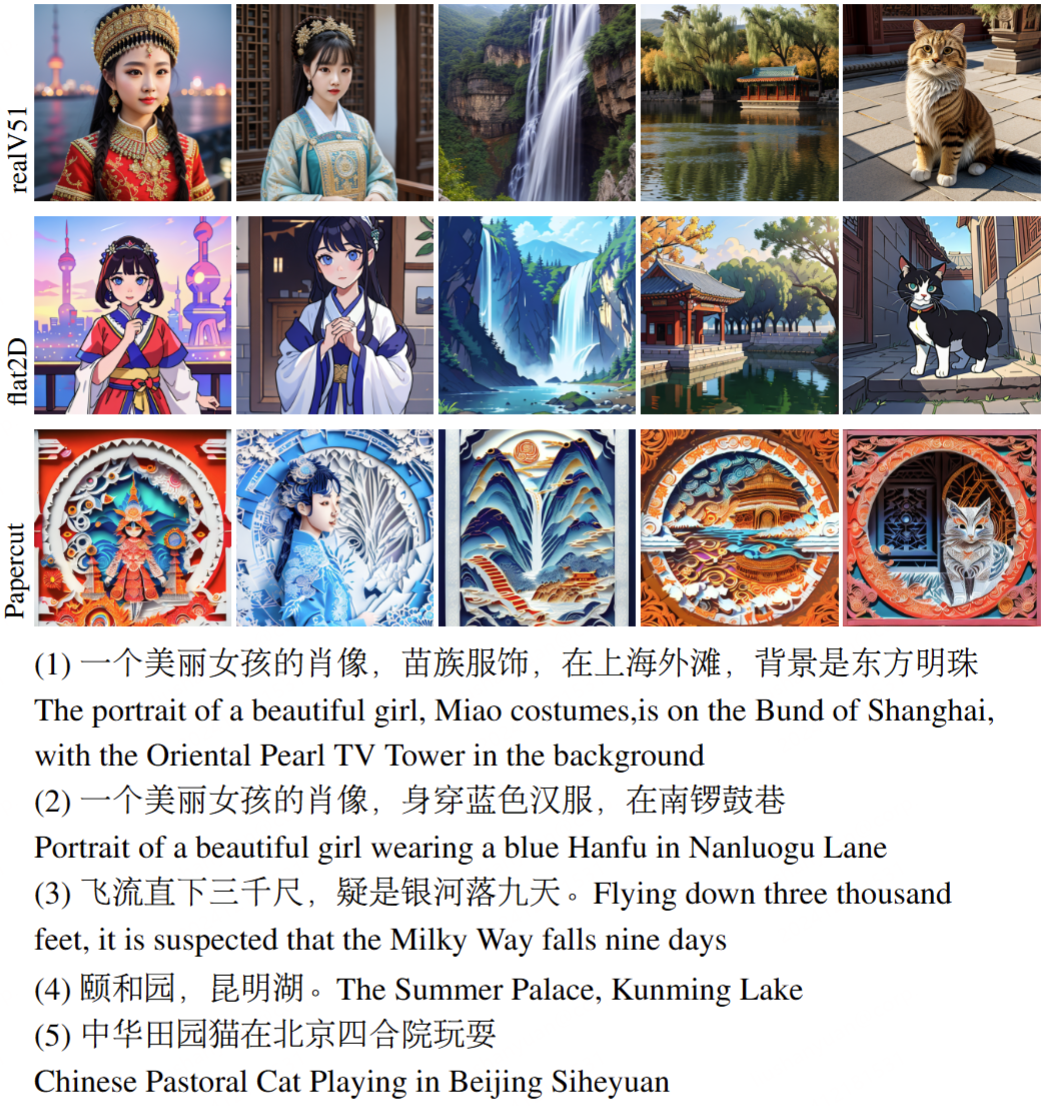} 
\caption{Checkpoint.}
\label{fig:style transfer Checkpoint}
\vspace{-1em}
\end{figure}

\begin{figure}[t]
\centering
\includegraphics[width=0.7\columnwidth]{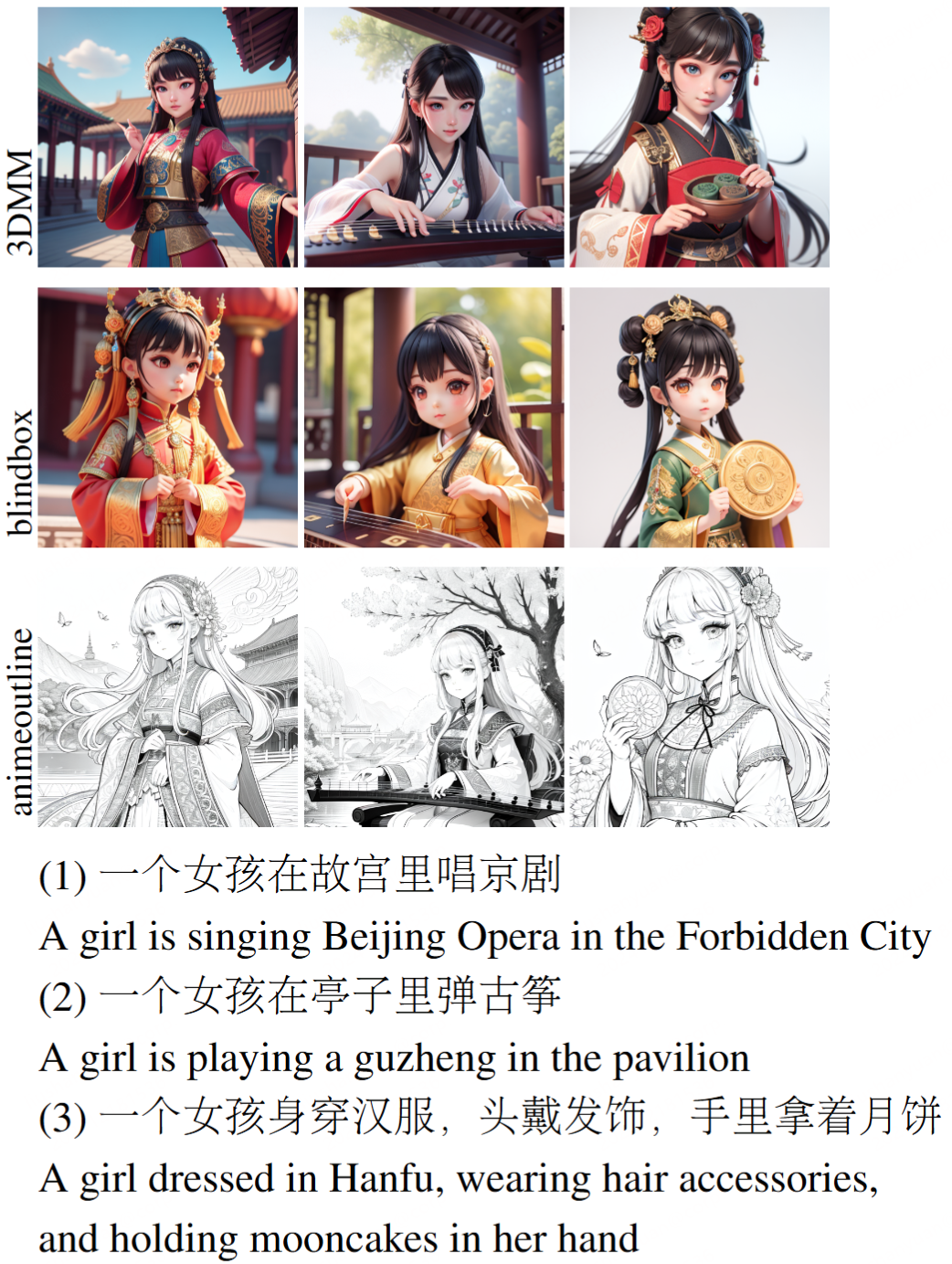} 
\caption{LoRA.}
\label{fig:LoRA}
\end{figure}

\begin{figure}[t]
\centering
\includegraphics[width=0.9\columnwidth]{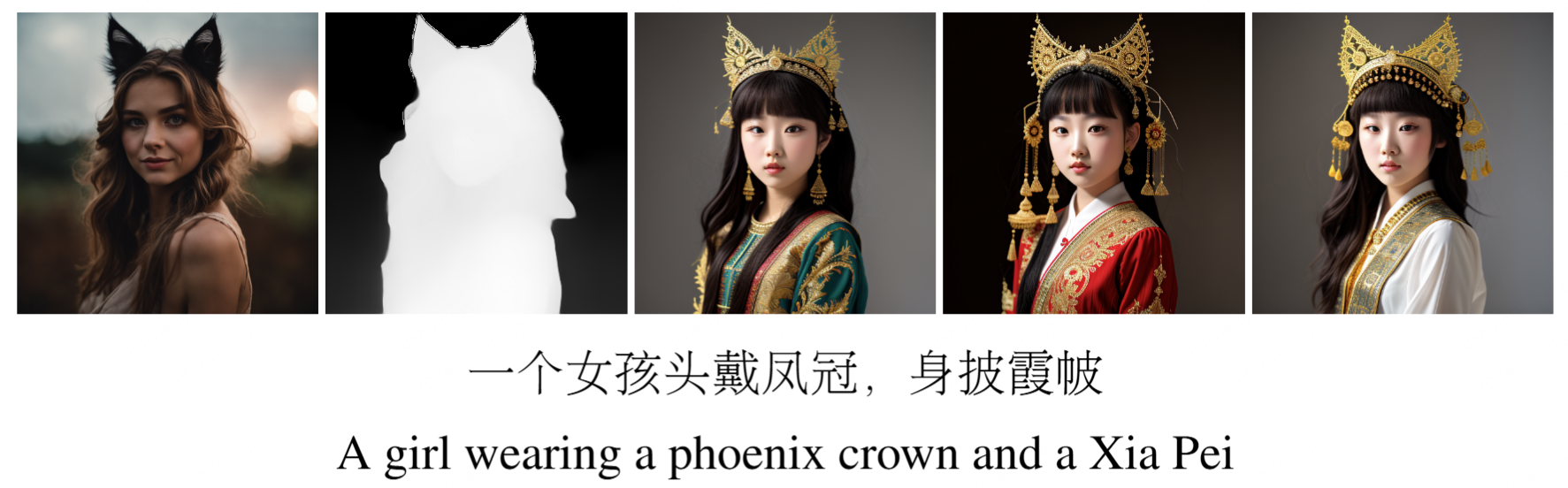} 
\caption{ControlNet.}
\label{fig:ControlNet}
\vspace{-1em}
\end{figure}

\begin{figure}[t]
\centering
\includegraphics[width=0.9\columnwidth]{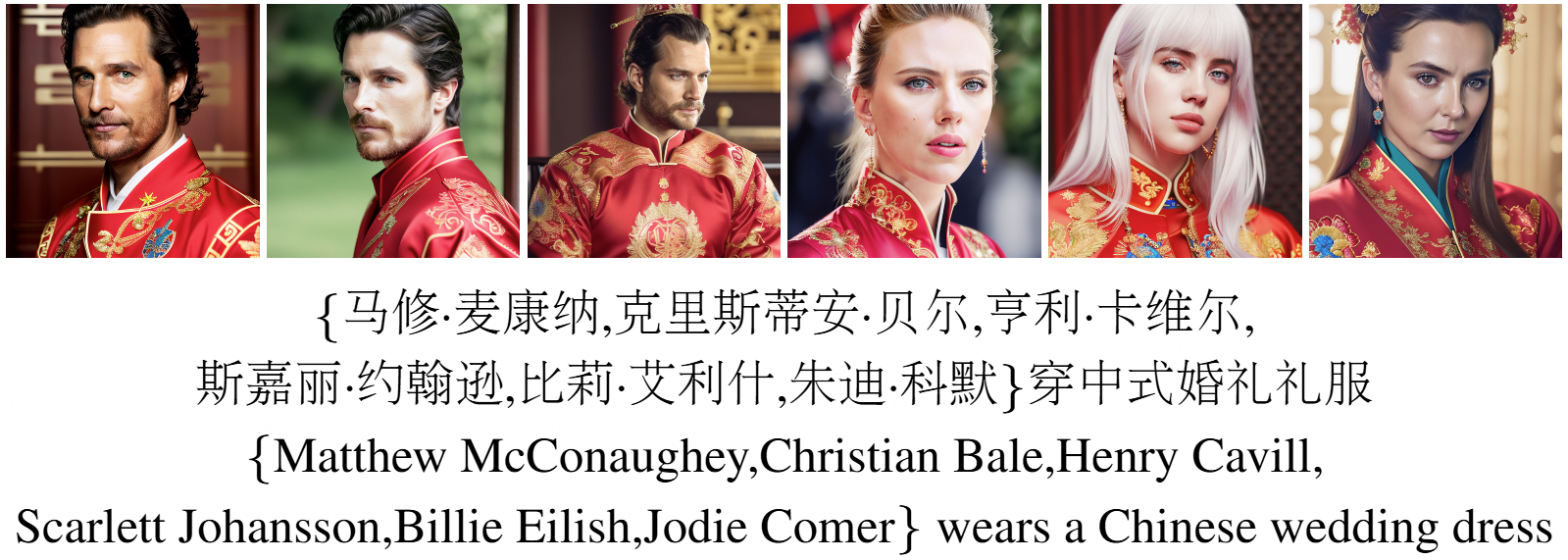} 
\caption{Dreambooth.}
\label{fig:DreamBooth}
\end{figure}

\begin{figure}[t]
\centering
\includegraphics[width=0.9\columnwidth]{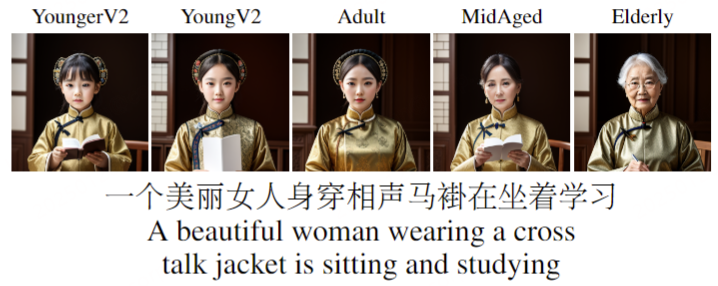} 
\caption{Textual Inversion.}
\label{fig:Text Inversion}
\vspace{-1em}
\end{figure}

\begin{figure}[t]
\centering
\includegraphics[width=0.9\columnwidth]{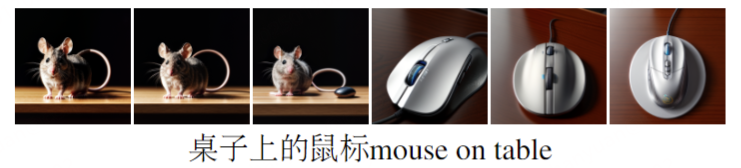} 
\caption{Ablation Study.}
\label{fig:Ablation}
\vspace{-1em}
\end{figure}

This section offers an overview of the experimental setup and showcases the effectiveness of BDM through both qualitative and quantitative demonstrations.
\subsection{Experimental Setup}
The training dataset comprises around one billion image-text pairs, which includes various internal Chinese datasets and parts of publicly available English datasets. To ensure the generation of meaningful Chinese concepts and minimize the impact of English concepts, our dataset is predominantly composed of Chinese data. We exclude data with watermarks, low aesthetic quality, or irrelevant image-text content.

Utilizing the latent space of Stable Diffusion 1.5\cite{Rombach_Blattmann_Lorenz_Esser_Ommer_2022}, BDM employs the same Variational Autoencoder (VAE) to facilitate transformations between images in pixel space and the latent space. The entire model is built using PyTorch and we use the AdamW\cite{DBLP:conf/iclr/LoshchilovH19} optimizer for training, setting a learning rate of 1e-5 and a batch size of 3200. The training process spans two months on 80 NVIDIA A800 GPUs.

\subsection{Quantitative Evaluation}
\noindent\textbf{Human Evaluation on BDM-870.}
We collected 870 diverse Chinese prompts from real users as benchmark for human evaluation of BDM model, and named it BDM-870. The BDM-870 are intentionally diversified, containing 25 categories. These 25 categories are evenly distributed, and more detailed data content is available in this link\footnote{\url{https://github.com/360CVGroup/Bridge_Diffusion_Model}}. Based on BDM-870, we conducted back-to-back evaluations on image quality and text-image alignment, at 512×512 resolution, against recent state-of-the-art models, such as ERNIE-ViLG 2.0 and Vega AI\footnote{Vega AI is a leading Chinese text-to-image creation platform.}\cite{vegaai}. Seven evaluators were presented with two sets of images generated by BDM and the competing model, but unaware of which. They are required to pick out the better image with respect to either image quality or text-image alignment. For image quality, aesthetics, object integrity, rationality of details such as face, fingers and limbs are mainly considered. Results are illustrated in Fig.\ref{fig:human}, and human raters prefer BDM over ERNIE-ViLG 2.0 by 54.57\%±3.1\% and 62.65\%±2.9\% and over Vega AI by 54.62\%±3.1\% and 58.79\%±3.0\% quantitatively. For Chinese-native text-to-image generation, BDM outperformed all other models considered in human evaluation.

\noindent\textbf{Evaluation on COCO.}
In line with previous research\cite{saharia2022photorealistic,feng2023ernie,balaji2022ediffi,Rombach_Blattmann_Lorenz_Esser_Ommer_2022}, we conducted an evaluation of BDM on the COCO 256 × 256 dataset using the zero-shot Frechet Inception Distance (FID), a metric that assesses image quality and diversity. We randomly selected 30,000 images from the validation set for assessment and translate the English captions into Chinese automatically. During training, BDM predominantly employs Chinese text-image data, resulting in a significant disparity between the world concepts in the output of BDM and COCO. Furthermore, there is a loss of accuracy in the process of automatically translating COCO captions into Chinese. In this context, BDM still manages to achieve FID of 9.93, maintaining a performance that closely aligns with SD1.5's 8.59. It even outperforms some English models, as shown in Table\ref{tab:Comparison}. This illustrates that BDM can effectively preserve the performance of English backbone with minimal loss while accommodating the generation of Chinese concepts. Moreover, to demonstrate the effectiveness of BDM's structure and strategies, no additional optimizations were applied to enhance the generation quality for both the English backbone and Chinese language branch. This highlights BDM's ability to easily and effectively realize a non-English language native model that is compatible with the English-speaking community.

\noindent\textbf{Chinese cultural inclination.}
To validate the Chinese cultural inclination of BDM, we generated 25 general prompts about concepts such as race, traditional architecture, food, and festivals using GPT. Then, we used BDM(RealisticVisionV51) and RealisticVisionV51 to generate 750 images. We calculated the CLIP score between the generated images and the concepts of "Chinese" and "Caucasian" to measure the model's cultural inclination, results are present in Table \ref{tab:clip score}. The table clearly shows that the CLIP score between BDM (RealV51) and "Chinese" consistently surpasses the CLIP score between BDM (RealV51) and "Caucasian", suggesting a stronger inclination towards generating Chinese cultural semantics. Conversely, the English model exhibits the opposite trend. Furthermore, the CLIP score between BDM (RealV51) and "Chinese" surpasses that between RealV51 and "Chinese", while the opposite trend is observed for "Caucasian", indicating that BDM demonstrates a stronger inclination towards Chinese cultural context compared to the English model.

\noindent\textbf{Training data scale.}
To show how the performance of BDM scales with data size, we present aesthetic metric records along BDM's training, as shown in Fig.\ref{fig:dataset scale}. BDM's performance continuously improves with more training data.

\subsection{Qualitative Results}
In this section, the semantic information for all images comes exclusively from the Chinese language branch which means that the English backbone receives only general descriptions such as "high definition", style descriptors like "CG", "anime", or trigger words, lacking any significant semantic input. All the plugins are obtained from Civitai.

\noindent\textbf{Native language semantics.}
In Fig.\ref{fig:native language semantic Checkpoint}, we demonstrate BDM's capability to generate native language semantic images. The original results of SD1.5 and the two versions of BDM are showcased: BDM(SD1.5) with Stable Diffusion 1.5\cite{Rombach_Blattmann_Lorenz_Esser_Ommer_2022} as the backbone, and BDM(RealisticVisionV51) with RealisticVisionV51\cite{realisticVisionV51} as the backbone. As shown by the results of prompts 1-3, the Chinese semantics generated by the English model are all incorrect, while BDM can generate accurately. Additionally, as shown by the results of prompts 4-5, due to the inherent translation issues of polysemy, the English model cannot generate the correct target, while BDM can avoid the impact of translation.

Moreover, it's worth mentioning that the image quality of the generated images by RealisticVisionV51 is significantly better than that of Stable Diffusion 1.5, despite BDM being trained on the latter. This observation emphasizes the idea that as the performance of the English backbone improves, BDM also sees improvements. Consequently, options such as refining the backbone or incorporating a more advanced English backbone can be considered to further enhance BDM's capabilities.

\noindent\textbf{Checkpoint.}
In Fig.\ref{fig:style transfer Checkpoint}, we present BDM's capability in effectively integrating with various English communities checkpoints. The backbone is alternately set to realisticVisionV51\cite{realisticVisionV51}(realV51 for short),flat2DAnimerge\cite{flat2DAnimerge}(flat2D for short) and midjourneyPapercut\cite{midjourneyPapercut}(Papercut for short). By using the same native language text as input across all configurations, it becomes evident that BDM, with various English backbones, can generate images that are not only semantically consistent within the native language context but also accurately reflect backbone's unique styles.

\noindent\textbf{LoRA.}
In Fig.\ref{fig:LoRA}, we demonstrate BDM's ability to smoothly integrate with LoRA\cite{Hu_Shen_Wallis_Allen-Zhu_Li_Wang_Chen_2021} within English communities. We select three different variants of LoRA, namely 3DMM\cite{3DMM}, blindbox\cite{blindbox}, and animeoutline\cite{animeoutline}, and find that BDM, in combination with each variant, is capable of generating Chinese semantic images that align with the respective styles of LoRA.

\noindent\textbf{ControlNet.}
In Fig.\ref{fig:ControlNet}, We showcase BDM's capability to work with ControlNet\cite{zhang2023adding}. We select \cite{controlDepth} as the control model and use depth maps to generate controlled images. The images produced feature individuals of East Asian descent, aligning with the distribution of BDM's training data, and their attire distinctly reflects Chinese elements.

\noindent\textbf{Dreambooth.}
In Fig.\ref{fig:DreamBooth}, we demonstrate BDM's successful integration with Dreambooth\cite{ruiz2023dreambooth}. We chose Dreambooth models representing six well-known figures from the Famous People checkpoint\cite{FamousPeople} and then produce images depicting these figures in traditional Chinese wedding attire. Clearly, the generated images accurately portray the chosen individuals, seamlessly incorporating relevant Chinese cultural elements.

\noindent\textbf{Textual Inversion.}
In Fig.\ref{fig:Text Inversion}, we showcase BDM's ability to work with Textual Inversion\cite{DBLP:conf/iclr/GalAAPBCC23}. We select English Textual Inversion embeddings capable of adjusting age, named Age Slider\cite{AgeSlider}, and carry out image generation using the same random seed and identical native language text descriptions. It is clear that within the native language context, the age of the depicted individuals can be varied by adjusting the embeddings.

\subsection{Ablation Study}
BDM's structure can be considered as a diffusion UNet with two different language encoders and one shared decoder. As mentioned in \ref{inference strategy}, by controlling text prompt input to different language branch, we can generate images emphasizing different language semantics. To further study BDM's capability in capturing different language's semantics, we conduct experiments by weighting the Chinese language branch, similar to the experiments on weighting the branches in ControlNet\cite{zhang2023adding} and LoRA\cite{Hu_Shen_Wallis_Allen-Zhu_Li_Wang_Chen_2021}. The weight increases from 0 to 1 with step 0.2. The Chinese language branch receives Chinese prompt meanwhile the English backbone receives corresponding English prompt (translated from Chinese prompt) as text input. Random seed is kept fixed during the experiment. We use realisticVisionV51\cite{realisticVisionV51} as the backbone model.

To amplify the study effect, we deliberately choose a case where the translated English prompt contains polysemous word "mouse". The Chinese prompt means "computer mouse on table" whereas the translated English prompt "mouse on table" can be considered as an animal mouse. As shown in Fig.\ref{fig:Ablation}, as the weight increases for BDM's Chinese language branch, the generated image gradually changes from an animal mouse to a computer mouse, aligning more closely with the Chinese semantics from BDM's Chinese language branch.

\section{Conclusions}
\label{sec:Conclusions}
We introduce the "Bridge Diffusion Model" (BDM) as a new structure to develop Chinese native TTI model which keeps compatibility with the English TTI communities. The English backbone within BDM is kept frozen thus maintains compatibility with its English ancestor model, meanwhile the non-English language branch is responsible for expressing native language meanings. Experiments show BDM can produce high-quality images from Chinese prompts and easily integrate with English TTI plugins.

\bibliography{aaai25}

\end{document}